# Backbone Fragility and the Local Search Cost Peak

**Josh Singer**                                                                JOSHUAS@DAI.ED.AC.UK
*Division of Informatics, University of Edinburgh*
*80 South Bridge, Edinburgh EH1 1HN, United Kingdom*

**Ian P. Gent**                                                                IPG@DCS.ST-AND.AC.UK
*School of Computer Science, University of St. Andrews*
*North Haugh, St. Andrews, Fife KY16 9SS, United Kingdom*

**Alan Smaill**                                                                A.SMAILL@ED.AC.UK
*Division of Informatics, University of Edinburgh*
*80 South Bridge, Edinburgh EH1 1HN, United Kingdom*

## Abstract

The local search algorithm WSAT is one of the most successful algorithms for solving the satisfiability (SAT) problem. It is notably effective at solving hard Random 3-SAT instances near the so-called 'satisfiability threshold', but still shows a peak in search cost near the threshold and large variations in cost over different instances. We make a number of significant contributions to the analysis of WSAT on high-cost random instances, using the recently-introduced concept of the *backbone* of a SAT instance. The backbone is the set of literals which are entailed by an instance. We find that the number of solutions predicts the cost well for small-backbone instances but is much less relevant for the large-backbone instances which appear near the threshold and dominate in the overconstrained region. We show a very strong correlation between search cost and the Hamming distance to the nearest solution early in WSAT's search. This pattern leads us to introduce a measure of the *backbone fragility* of an instance, which indicates how persistent the backbone is as clauses are removed. We propose that high-cost random instances for local search are those with very large backbones which are also backbone-fragile. We suggest that the decay in cost beyond the satisfiability threshold is due to increasing backbone robustness (the opposite of backbone fragility). Our hypothesis makes three correct predictions. First, that the backbone robustness of an instance is negatively correlated with the local search cost when other factors are controlled for. Second, that backbone-minimal instances (which are 3-SAT instances altered so as to be more backbone-fragile) are unusually hard for WSAT. Third, that the clauses most often unsatisfied during search are those whose deletion has the most effect on the backbone. In understanding the pathologies of local search methods, we hope to contribute to the development of new and better techniques.

## 1. Introduction

Why do some problem instances require such a high computational cost for algorithms to solve? Answering this question will help us to understand the interaction between search algorithms and problem instance structure and can potentially suggest principled improvements, for example the Minimise-Kappa heuristic (Gent, MacIntyre, Prosser, & Walsh, 1996; Walsh, 1998).

In this paper we study the propositional satisfiability problem (SAT). SAT is important as it was the first and is perhaps the archetypal NP-complete problem. Furthermore, many

AI tasks of practical interest such as constraint satisfaction, planning and timetabling can be naturally encoded as SAT instances.

A SAT instance $C$ is a propositional formula in conjunctive normal form. $C$ is a bag of $m$ clauses which represents their conjunction. A *clause* is a disjunction of *literals*, which are Boolean variables or their negations. The variables constitute a set of $n$ symbols $V$. An *assignment* is a mapping from $V$ to {true, false}. The decision question for SAT asks whether there exists an assignment which makes $C$ true under the standard logical interpretation of the connectives. Such an assignment is a *solution* of the instance. If there is a solution, the SAT instance is said to be *satisfiable*. In this study, assignments where a few clauses are unsatisfied are also important. We refer to these as *quasi-solutions*. $k$-SAT is the SAT problem restricted to clauses containing $k$ literals. Notably, the $k$-SAT decision problem is NP-hard for $k \geq 3$ (Cook, 1971). In several NP-hard decision problems, such as 3-SAT, certain probabilistic distributions of instances parameterised by a "control parameter" exhibit a sharp threshold or "phase transition" in the probability of there being a solution (Cheeseman, Kanefsky, & Taylor, 1991; Gent et al., 1996; Mitchell, Selman, & Levesque, 1992). There is a critical value of the control parameter such that instances generated with the parameter in the region lower than the critical value (the *underconstrained* region) almost always have solutions. Those generated from the *overconstrained* region where the control parameter is higher than the critical value almost always have no solutions.

In many problem distributions, this threshold is associated with a peak in search cost for a wide range of algorithms. Instances generated from the distribution with the control parameter near the critical value are hardest and cost decays as we move from this value to lower or higher values. This behaviour is of interest to basic AI research. Being devoid of any regularities, random instances represent the challenge faced by an algorithm in the absence of any assumptions about the problem domain, or once all knowledge about it has been exploited in the design of the algorithm or transformation of the problem instance.

Random $k$-SAT is a parameterised distribution of $k$-SAT instances. In Random $k$-SAT, $n$ is fixed and the control parameter is $m/n$. Varying $m/n$ produces a sharp threshold in the probability of satisfiability and an associated cost peak for a range of complete algorithms (Crawford & Auton, 1996; Larrabee & Tsuji, 1992). The cost peak pattern in Random $k$-SAT has been conjectured to extend to all reasonable complete methods by Cook and Mitchell, (1997) who also give an overview of analytic and experimental results on the average-case hardness of SAT distributions.

In this paper we study the behaviour of *local search* on Random $k$-SAT. The term local search encompasses a class of procedures in which a series of assignments are examined with the objective of finding a solution. The first assignment is typically selected at random. Local search then proceeds by moving from one assignment to another by "flipping" (i.e. inverting) the truth value of a single variable. The variable to flip is chosen using a heuristic which may include randomness, an element of hill-climbing (for example on the number of satisfied clauses) and memory. Usually, local search is incomplete for the SAT decision problem: there is no guarantee that if a solution exists, it will be found within any time bound. Unlike complete procedures, local search cannot usually prove for certain that no solution exists.

It is a relatively recent discovery (e.g. Selman, Levesque and Mitchell, 1992) that the average cost for local search procedures scales much better than that of the best complete procedures at the critical value of $m/n$ in Random 3-SAT. More recent studies, (e.g. Parkes and Walser, 1996) have confirmed this in detail. Therefore in any system where completeness may be sacrificed, local search procedures have an important role to play, and this is why they have generated so much interest in recent years.

If we restrict ourselves to those instances of the distribution which are satisfiable and increase the control parameter, there is a peak in the cost for local search procedures to solve the instances near the critical value in several constraint-like problems (Clark, Frank, Gent, MacIntyre, Tomov, & Walsh, 1996; Hogg & Williams, 1994). In the underconstrained region, the average cost increases with $m/n$ due to the decreasing number of solutions per instance (Clark et al., 1996). However, in the overconstrained region, the cost decreases with $m/n$ although the number of solutions per instance continues to fall. Several researchers have noted this fact with surprise (Clark et al., 1996; Parkes, 1997; Yokoo, 1997) since the number of solutions does not change in any special way near the critical value. Why, then, should the cost of satisfiable instances peak near the critical value, and then decay?

Parkes (1997) provided an appealing answer to the first part of this question in his study of the *backbone* of satisfiable Random 3-SAT instances. For satisfiable SAT instances, the backbone is the set of literals which are logically entailed by the clauses of the instance[1]. Variables which appear in these entailed literals are each forced to take a particular truth value in all solutions. Parkes' study demonstrated that in instances from the underconstrained region, only a small fraction of the variables, if any, appear in the backbone. However, as the control parameter is increased towards the critical value, a subclass of instances which have large backbones, mentioning around 75-95% of the variables, rapidly emerges. Soon after the control parameter is increased into the overconstrained region these large-backbone instances account for all but a few of the satisfiable instances. Parkes also showed that for a fixed value of the control parameter, the cost for the local search procedure WSAT is strongly influenced by the size of the backbone. This suggests that the peak in average WSAT cost near the critical value as the control parameter is increased may be due to the emergence of large-backbone instances at this point. Parkes noted that for any given size of backbone, the cost is actually higher for instances from the underconstrained region and falls as the control parameter is increased. He also identified this fall as indicative of another factor which produces the overall peak in cost. The main aim of this paper is to identify the factor responsible for this pattern; why are some instances with a certain size of backbone more costly to solve than others?

The remainder of the paper is organised as follows. In Section 2 we review the details of the WSAT algorithm and the Random $k$-SAT distribution and discuss the experimental conditions which were used. We also elucidate the patterns in cost which we intend to explain and show how the number of solutions and the backbone size interact. In Section 3 we identify a remarkable pattern in WSAT's search behaviour which clearly distinguishes high cost from lower cost instances of a certain backbone size. WSAT is usually drawn early on in the search to quasi-solutions where a few clauses are unsatisfied. On high cost instances, these quasi-solutions are distant from the nearest solution, while on lower cost instances of equal backbone size, they are less distant. In Section 4 we develop a causal hypothesis, postulating a structural property of instances which induces a search space structure which in turn causes the observed search behaviour and thus the cost pattern. We suggest that instances of a certain backbone size are of high cost when they are *backbone-fragile*, i.e. when the removal of a few clauses at random results in an instance with a greatly reduced backbone size. We discuss how this property may be measured and show how as the control parameter is increased, instances of a certain backbone size become less backbone-fragile.

---

1. Here, our use of the term "backbone" follows Monasson, Zecchina, Kirkpatrick, Selman and Troyansky (1999a, 1999b) whose definition of the backbone is equivalent to ours for satisfiable instances.

A hypothesis is only of true scientific merit if it makes correct predictions. Our hypothesis made three correct predictions for which we provide experimental evidence. In Section 5 we show that the degree to which an instance is backbone-fragile accounts for some of the variance in cost when the control parameter and the backbone-size are fixed. In Section 6 we consider the generation of instances which are very backbone-fragile. If clauses are removed such that the backbone is unaffected, we found that the resulting instances became progressively more backbone-fragile. Eventually, no more clauses can be removed without affecting the backbone and the instance is said to be *backbone minimal*. Our hypothesis correctly predicts that as clauses are removed in this way from Random 3-SAT instances, the cost becomes considerably higher. In Section 7 we show that the hypothesis makes a correct prediction relating to the search behaviour: the clauses which are most often unsatisfied during search are those whose removal most affects the backbone. In Section 8 we relate this study to previous research and give suggestions for further work. Finally, Section 9 concludes.

## 2. Background

In this section we discuss the local search algorithm WSAT, the measurement of computational cost for it and its representativeness of local search algorithms in general. We also review the Random $k$-SAT distribution and the overall cost pattern for WSAT on Random $k$-SAT. Finally we look at how backbone size and the number of solutions interact to affect the cost.

### 2.1 The WSAT Algorithm

The term WSAT was first introduced by Selman *et al.* (1994). It refers to a local search architecture which has also been the subject of a number of subsequent empirical studies (Hoos, 1999a; McAllester, Selman, & Kautz, 1997; Parkes & Walser, 1996; Parkes, 1997). A pseudocode outline of the WSAT algorithm is given in Figure 1. An important feature of WSAT is that, unlike earlier local search algorithms, it chooses an unsatisfied clause and then flips a variable appearing in that clause: SELECT-VARIABLE-FROM-CLAUSE must return a variable mentioned in *clause*. This architecture was first seen in the "random walk algorithm" due to Papadimitriou (1991). WSAT may use different strategies for SELECT-VARIABLE-FROM-CLAUSE. In this study, we used the "SKC" strategy introduced by Selman, Kautz and Cohen (1994); we refer to this combination simply as WSAT. Pseudocode for the SKC strategy is given in Figure 2.

We follow Hoos (1998) in our approach to measuring the computational cost of SAT instances for our local search algorithm WSAT. Rather than run-times, we measure run-lengths : the number of flips taken to find a solution. We set the noise level $p$ to 0.55, which Hoos found to be approximately optimal on Random 3-SAT. Hoos and Stützle (1998) showed

WSat($C$, Max-tries, Max-flips, $p$)
    **for** $i = 1$ to Max-tries
        $T :=$ a random assignment
        **for** $j = 1$ to Max-flips
            $clause :=$ an unsatisfied clause of $C$, selected at random
            $v :=$ Select-variable-from-clause($clause$, $C$, $p$)
            $T := T$ with $v$'s value "flipped"
            **if** $T$ is satisfying
                return $T$
            **end if**
        **end for**
    **end for**
    **return** "no satisfying assignment found"

Figure 1: The WSat local search algorithm

Select-variable-from-clause($clause$, $C$, $p$)
    **for** each variable $x$ mentioned in $clause$
        $breaks[x] :=$ the number of clauses in $C$ which would
                become unsatisfied if $x$ were flipped
    **end for**
    **if** there is some variable $y$ from $clause$ such that $breaks[y] = 0$
        **return** such a variable, breaking ties randomly
    **else**
        with probability $1 - p$
            **return** a variable $z$ from $clause$
                which minimises $breaks[z]$, breaking ties randomly
        with probability $p$
            **return** a variable $z$ from $clause$
                chosen randomly
    **end if**

Figure 2: The SKC variable selection strategy

that run lengths on all but the easiest instances are exponentially distributed for many local search variants. This implies that the random "restart" mechanism (the re-randomisation of $T$ after Max-flips flips) is not significantly worthwhile.

It is not known to date whether, without using restart, WSAT will almost surely (i.e. with probability approaching 1) find a solution on satisfiable 3-SAT instances if given unlimited flips. If a local search algorithm will eventually find a solution under these conditions, it is said to be probabilistically approximately complete (PAC). Hoos (1999a) proved whether several local search algorithms were PAC and Culberson and Gent (1999a) proved that WSAT is PAC for the 2-SAT case. Hoos (1998) observed that his data suggested WSAT could be PAC. We set Max-tries to 1 and Max-flips infinite on all runs reported in this paper. A solution was found in every run, which is further evidence that WSAT may be PAC.

Another implication of the exponential distribution of run lengths is that a large number of samples must be taken to obtain a good estimate of the mean. Following Hoos, we use the median of 1000 WSAT runs on each instance as our descriptive statistic representing WSAT's search cost on that instance. This appears to give a stable estimate of the cost (as it is less sensitive to the long tail than the mean) with only a moderate amount of computational effort.

One objection to studying a single algorithm from the local search class is that it may not be representative: results obtained for the algorithm may not generalise to other members of the class. While we accept this objection, there is evidence that under certain conditions, one local search algorithm is actually to a large extent representative of the whole class. For example Hoos (1998) found a very high correlation between the computational costs of random instances of pairs of different local search algorithms, including WSAT. This also suggests that there is some algorithm-independent property of these instances which results in high cost for this class of algorithms.

## 2.2 Random $k$-SAT

We use the well-studied Random $k$-SAT distribution (Franco & Paull, 1983; Mitchell et al., 1992) with $k = 3$. Random $k$-SAT is a distribution of $k$-SAT instances, parameterised by the ratio of clauses to variables $m/n$. Let $V$ be the fixed set of Boolean variable symbols of size $n$. To generate an instance from Random $k$-SAT with $m$ clauses and $n$ variables, each clause in $C$ is independently chosen by randomly selecting as its literals $k$ distinct variables from $V$ and independently negating each with probability $\frac{1}{2}$. There is no guarantee that all variables are mentioned in the instance or that it will not contain duplicate clauses.

As local search cannot solve unsatisfiable instances, we filter these out using a complete SAT procedure. In order to control for the effects of the backbone size, we will also need to isolate the portion of the satisfiable part of the distribution for which the backbone size is of a certain size. This is obtained by calculating the backbone size of each satisfiable instance and discarding those whose backbone is not of the required size. We will term this *controlling* the backbone size. Satisfiable instances with certain backbone sizes are rare at certain values of $m/n$. For example when $m/n$ is 4.49, we found that only 1 in about 20,000 generated instances is satisfiable with a backbone size of 10. Hence generation of instances in this way can be somewhat costly in computational terms. This was therefore one of the limits on the value of $n$ for which data could be collected.

We were primarily interested in the threshold region of the control parameter, where the cost peak occurs: the region near the point at which 50% of the instances are satisfiable. We looked at the region between 90% and 20% satisfiability.

### 2.3 A Pattern in WSat Cost for Random 3-SAT

In Figure 3 we show the peak in WSat cost which has been mentioned e.g. by Parkes (1997). The peak is slightly above the 50% point (4.29) for the median but appears to shift down for higher percentiles. A similar pattern was noticed by Hogg and Williams (1994) in local search cost for graph colouring.

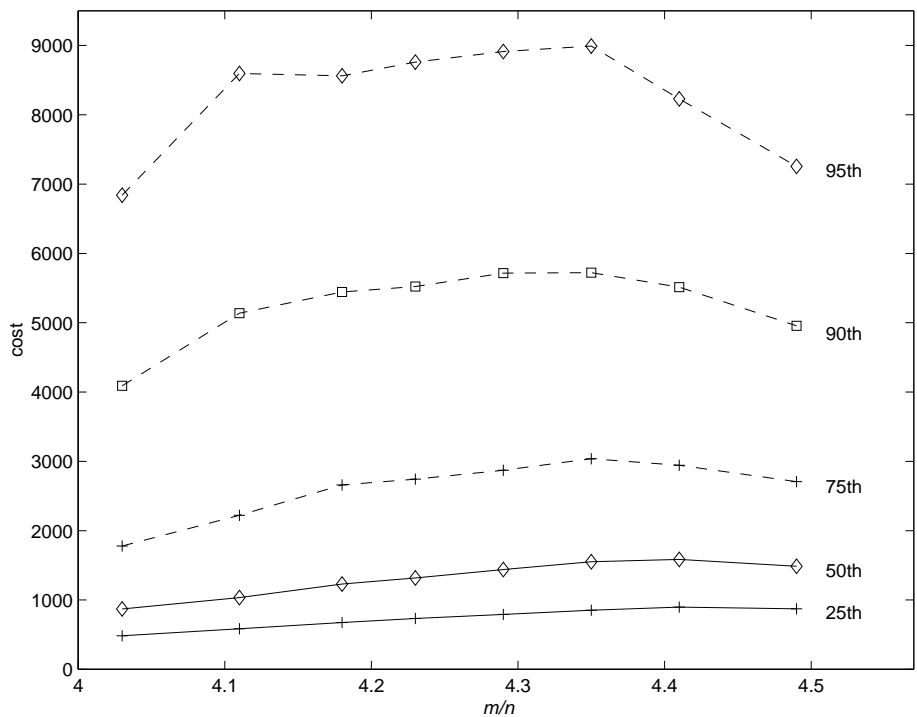

Figure 3: The cost peak for WSat as $m/n$ is varied. At each level of $m/n$, we generated 5000 satisfiable instances. We measured per-instance WSat cost for each of these. Each line in the plot gives a different point in the cost distribution, e.g. the 90th percentile is the difficulty of the 500th most costly instance for WSat.

Both Parkes (1997) and Yokoo (1997) suggest that the local search cost peak shown for WSat in Figure 3 is a result of two competing factors. As $m/n$ is increased the number of solutions per instance falls and this causes the onset of high cost. However, the number of solutions continues to fall in the overconstrained region but the cost decreases. There must therefore be a second factor whose effect outweighs that of the number of solutions in the overconstrained region so as to cause the fall in cost. The main aim of this paper is to identify this factor. A pattern in WSat cost on Random 3-SAT identified by Parkes (1997) is our starting point. Parkes observed that for a given backbone size and $n$, the average cost falls as $m/n$ is increased.

Figure 4 shows the fall in WSAT cost for $n = 100$ Random 3-SAT instances. Each point in the plot is the median cost of 1000 instances[2] and the length of the bars is the interquartile range of instance cost. The fall in cost is an approximately exponential decay for a range of $m/n$ near the threshold and for a range of backbone sizes. The rate of decay is affected by the backbone size, with the cost of large-backbone instances decaying fastest. The length of the error bars in Figure 4 along with the log scale of the cost axis indicates that the distribution of per-instance cost is also positively skewed even once backbone size is controlled. For example at the point where $m/n$ is 4.11 and backbone size is $0.9n$ the difference between the 75th percentile and the median is about 4000 whereas between the median and the 25th percentile it is about half that. The spread of cost is large, particularly relative to the effect of the control parameter. We do not think that a significant portion of this variance in the cost among instances is due to errors in our estimates of the cost for each instance.

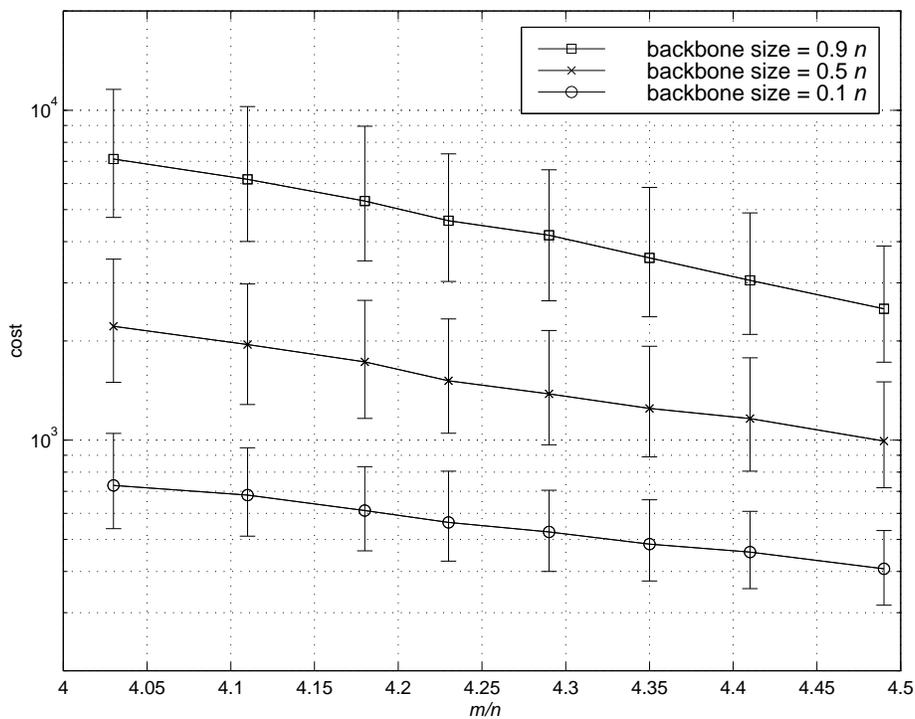

Figure 4: The effect of varying $m/n$ on cost while backbone size is controlled.

---

2. The cost of each instance is defined as the median run length of 1000 runs so each point in Figure 4 is a median of medians.

## 2.4 The Number of Solutions when Backbone Size is Controlled

We studied the effect of the number of solutions on WSAT cost. The number of solutions was determined using a modified complete procedure. For small-backbone instances, there was some evidence that the number of solutions actually increases with $m/n$, at least in the overconstrained region. Figure 5 shows a plot of the number of solutions, with backbone size controlled at $0.1n$. Each point is the median of 1000 instances and the bars show the interquartile range. This possible increase in the number of solutions may help to explain the fall in cost for small-backbone instances, but it appears to be too weak an effect to account for it in full.

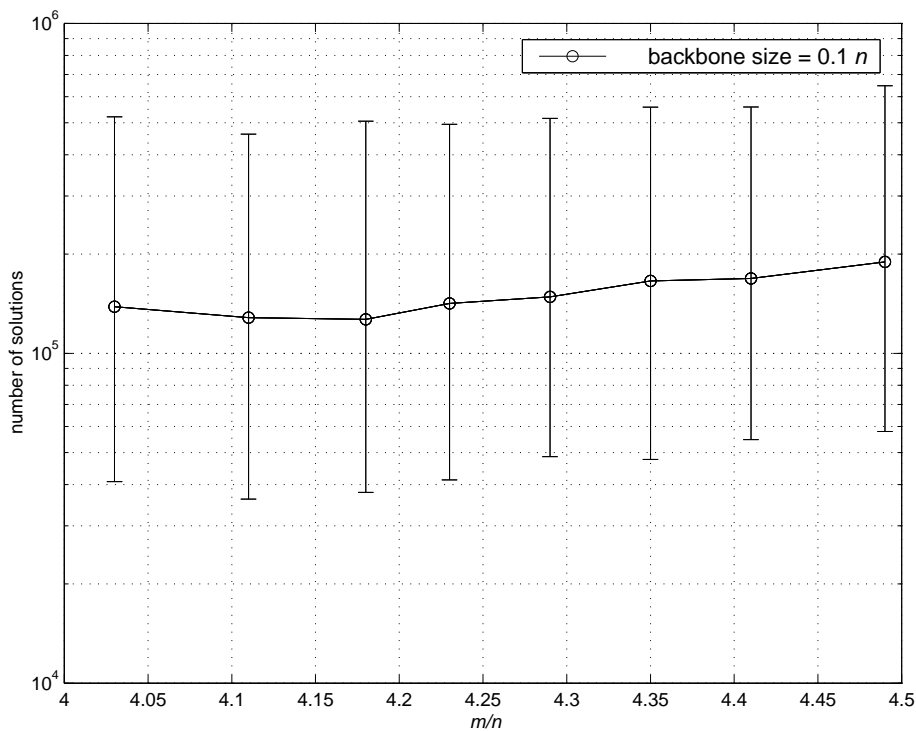

Figure 5: Number of solutions with $n = 100$, $m/n$ varied, and backbone size controlled at $0.1n$.

We studied the relationship between the number of solutions and the WSAT cost with backbone size controlled at different values. Figure 6 shows a log-log plot of the number of solutions against cost, where $m/n$ is 4.29 and backbone size is $0.1n$. A linear least squares regression (*lsr*) fit is superimposed. Table 1 gives summary data on the log-log scatter plot for different backbone sizes through the transition : the gradient and intercept of *lsr* fits, the product-moment correlation $r$ and the rank correlation.

The number of solutions is strongly and negatively related to the cost for smaller backbone sizes through the transition and the strength of the relationship is fairly constant as $m/n$ is varied. We speculate that the strong relationship on these instances arises because finding the backbone is straightforward and the main difficulty is encountering a solution once the backbone has been satisfied. The density of solutions in the region satisfying the backbone is then important. For larger backbone sizes, the number of solutions is less relevant to the cost. No significant change in the number of solutions for large backbone instances was observed as $m/n$ was varied. That the number of solutions and the cost are not strongly related for these instances is unsurprising, as the large backbone size implies that the solutions lie in a compact cluster and local search's main difficulty is finding this cluster (i.e. satisfying the backbone). Therefore we expect that the density of solutions within the cluster is not so important. Hoos (1998) observed that the correlation between number of solutions and local search cost becomes small in the overconstrained region. This can now be explained simply by the fact that the large-backbone instances dominate in this region.

| $m/n$ | Backbone size | Intercept of *lsr* fit | Gradient of *lsr* fit | $r$ | Rank corr. |
|---|---|---|---|---|---|
| 4.03 | $0.1n$ | 3.8993 | $-0.1967$ | $-0.7808$ | -0.7731 |
|      | $0.5n$ | 4.1410 | $-0.2123$ | $-0.6761$ | -0.6699 |
|      | $0.9n$ | 4.2070 | $-0.1372$ | $-0.1307$ | -0.1365 |
| 4.11 | $0.1n$ | 3.8727 | $-0.1989$ | $-0.7696$ | -0.7669 |
|      | $0.5n$ | 4.1551 | $-0.2304$ | $-0.6834$ | -0.6855 |
|      | $0.9n$ | 4.1387 | $-0.1336$ | $-0.1275$ | -0.1291 |
| 4.18 | $0.1n$ | 3.7867 | $-0.1911$ | $-0.7664$ | -0.7760 |
|      | $0.5n$ | 4.0533 | $-0.2180$ | $-0.6932$ | -0.6974 |
|      | $0.9n$ | 4.0202 | $-0.1146$ | $-0.1159$ | -0.1217 |
| 4.23 | $0.1n$ | 3.7771 | $-0.1932$ | $-0.7829$ | -0.7873 |
|      | $0.5n$ | 3.9890 | $-0.2140$ | $-0.6729$ | -0.6867 |
|      | $0.9n$ | 3.9891 | $-0.1270$ | $-0.1317$ | -0.1329 |
| 4.29 | $0.1n$ | 3.7309 | $-0.1910$ | $-0.7787$ | -0.7844 |
|      | $0.5n$ | 3.9169 | $-0.2076$ | $-0.6921$ | -0.6941 |
|      | $0.9n$ | 3.7836 | $-0.0610$ | $-0.0612$ | -0.0534 |
| 4.35 | $0.1n$ | 3.6981 | $-0.1896$ | $-0.8007$ | -0.7994 |
|      | $0.5n$ | 3.8933 | $-0.2133$ | $-0.6872$ | -0.6967 |
|      | $0.9n$ | 3.8173 | $-0.1018$ | $-0.1044$ | -0.0903 |
| 4.41 | $0.1n$ | 3.6083 | $-0.1782$ | $-0.7784$ | -0.7628 |
|      | $0.5n$ | 3.8445 | $-0.2094$ | $-0.7024$ | -0.7085 |
|      | $0.9n$ | 3.7772 | $-0.1120$ | $-0.1179$ | -0.1045 |
| 4.49 | $0.1n$ | 3.5483 | $-0.1748$ | $-0.7972$ | -0.7932 |
|      | $0.5n$ | 3.7577 | $-0.2043$ | $-0.6954$ | -0.6991 |
|      | $0.9n$ | 3.6228 | $-0.0842$ | $-0.0992$ | -0.0783 |

Table 1: Data on log-log correlations between number of solutions and cost with $n = 100$, $m/n$ varied and backbone size fixed at different values.

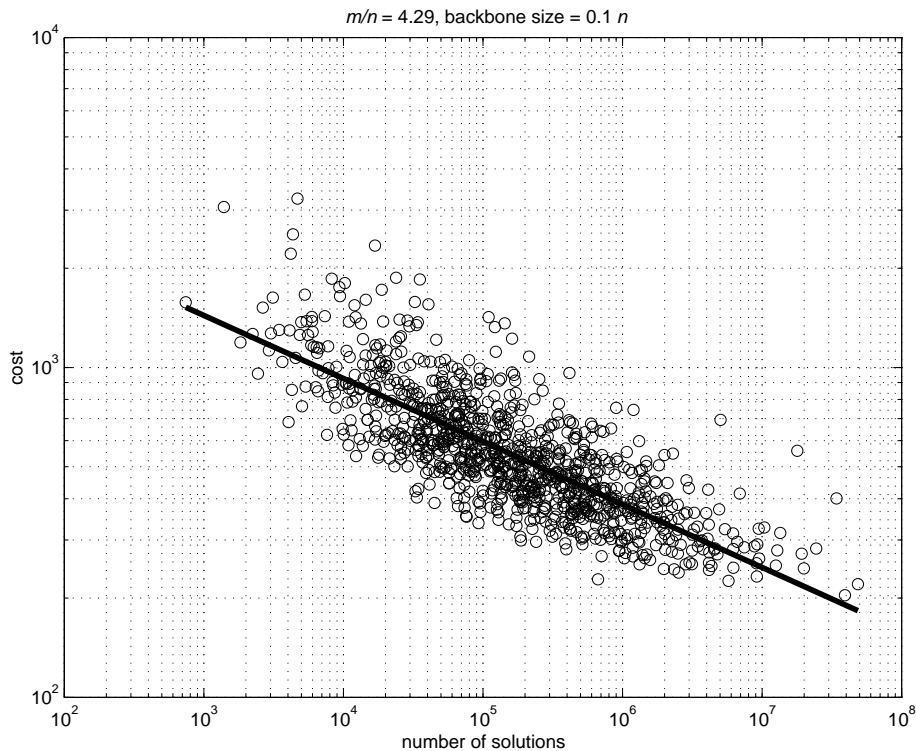

Figure 6: Scatter plot of number of solutions and cost with $n = 100$, $m/n = 4.29$ and backbone size fixed at $0.1n$.

## 3. Search Behaviour: the Hamming Distance to the Nearest Solution

In order to suggest the cause of the cost decay for large-backbone instances which was observed in Section 2.3, we made a detailed study of WSAT's *search behaviour*, i.e. the assignments visited during search. We report on this exploratory part of the research in this section. We explain the somewhat novel search behaviour metrics which were used before giving results and our discussion of them.

## 3.1 Definitions and Methods

Assuming a local search algorithm is PAC, in any given run of unlimited length, $f_b$, the number of flips taken to find the first assignment where at most $b$ clauses are unsatisfied, is well-defined for $b \geq 0$. $f_0$ is then equal to the run length.

A particular run of a local search algorithm then consists of a series of assignments $T_0, T_1, ..., T_{f_0}$, where $T_i$ is the assignment visited after $i$ flips have been made. We found that on Random 3-SAT with $n = 100$, an assignment satisfying all but a few clauses was quickly found and that during the remainder of the search, few clauses (1 - 10) were unsatisfied. As shown by Gent and Walsh (1993) in GSAT, there is a rapid hill-climbing phase, which is also suggested by Hoos (1998), followed by a long plateau-like phase in which the number of unsatisfied clauses is low but constantly changing. In our experiments we used $f_5$ as an arbitrary indicator of the length of the hill-climbing phase. Unlike in GSAT, in WSAT there is no well-defined end point for the hill-climbing phase, since short bursts of hill-climbing continue to occur for the rest of the search. We think that using $f_b$ as the indicator with any value of $b$ between 1 and 10 would give similar results.

Local search proceeds by flipping variable values and so we might expect that the Hamming distance between the current assignment and the nearest solution may also be of interest. The Hamming distance between two assignments $hd(T_1, T_2)$ is simply the number of variables which $T_1$ and $T_2$ assign differently. We studied the Hamming distance between the current assignment $T$ and the solution $T_{sol}$ of $C$ for which $hd(T, T_{sol})$ is minimised. We abbreviate this $hdns(T, C)$ (Hamming distance to nearest solution). For any assignment $T$, $hdns(T, C)$ may be calculated by using a complete SAT procedure which is modified so that every solution to $C$ is visited and its Hamming distance from $T$ calculated.

## 3.2 Results

In this section, data is based on Random 3-SAT instances with $n = 100$ and backbone size controlled at various values between $0.1n$ and $0.9n$. Recall that to control backbone at a certain value, we generate satisfiable Random 3-SAT instances as usual and discard all those whose backbone is not of the required size. We varied $m/n$ from the point of 90% satisfiability (4.03) to the point of 20% satisfiability (4.49). $hdns(T_{f_5}, C)$ is the Hamming distance between the first assignment at which no more than 5 clauses are unsatisfied and the nearest solution. For each instance we calculated the median value for $f_5$ and the mean value for $hdns(T_{f_5}, C)$ based on 1000 runs of WSAT. In the plots in Figures 7 and 8, each point is the median of 1000 instances.

Figure 7 shows the effect of varying $m/n$ on $f_5$ when backbone size is controlled. The values for $f_5$ are low compared to the cost and the range is very small. So although the cost to find a solution varies considerably from instance to instance, quasi-solutions are quickly found no matter what the overall cost. However, there are some notable effects of backbone size and $m/n$ on $f_5$. As might be expected, on the larger backbone instances, for which overall cost is generally higher, WSAT takes slightly longer to find a quasi-solution. The effect of $m/n$ is unexpected. If backbone size is controlled at $0.5n$ or more, as $m/n$ is increased WSAT takes slightly longer to find a quasi-solution, although simultaneously cost is decreasing as we have seen in Figure 4.

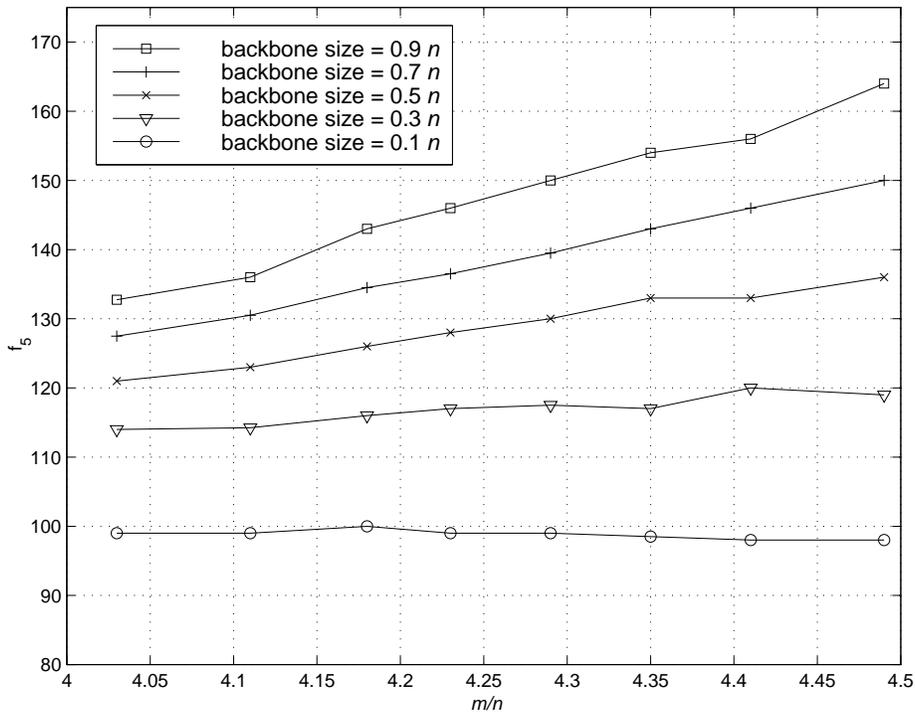

Figure 7: The effect of varying $m/n$ on $f_5$ while backbone size is controlled.

Figure 8 shows the effect of varying $m/n$ on $hdns(T_{f_5}, C)$ when the effects of backbone size are controlled for. In this plot, the bars give the interquartile range. The spread of values for mean $hdns(T_{f_5}, C)$ at each point is also small relative to the effect of varying $m/n$. Again the positive effect of backbone size on $hdns(T_{f_5}, C)$ is as one might expect since backbone size affects cost.

With backbone size controlled, as $m/n$ is increased through the satisfiability threshold, mean $hdns(T_{f_5}, C)$ decreases linearly for a wide range of backbone values. Hence, although a quasi-solution $(T_{f_5})$ is usually quickly found, on the instances of lower $m/n$ this quasi-solution is considerably Hamming-distant from the nearest solution. As $m/n$ is increased, while the backbone size is controlled, this effect is gradually lessened.

We also looked at the relationship between the search behaviour and the cost when $m/n$ was fixed and the backbone size was controlled. We found that in this case variance in $hdns(T_{f_5}, C)$ accounts for most of the cost variance. Figure 9 shows a plot of the mean $hdns(T_{f_5}, C)$ against the cost with backbone size controlled at $0.5n$ and $m/n$ fixed at 4.29. An $lsr$ fit is superimposed. The plot suggests $hdns(T_{f_5}, C)$ is linearly related to log of cost.

Table 2 gives the intercept and gradient for $lsr$ fits and $r$ values with backbone size controlled at three values and $m/n$ varied. Variance in $hdns(T_{f_5}, C)$ accounts for most of the variance in cost at three different backbone sizes and this is consistent through the threshold. The scatter plots (not shown) and linear *lsr* fits to the data were similar in shape to that of Figure 9 and so are consistent with a linear relationship. The $r$ values are greatest for small-backbone instances but the reasons for this are unclear. Possibly, since the search is shorter on the small-backbone instances, success follows quickly after $f_5$ and so $hdns(T_{f_5}, C)$ is a better indicator of the likelihood of finding a solution.

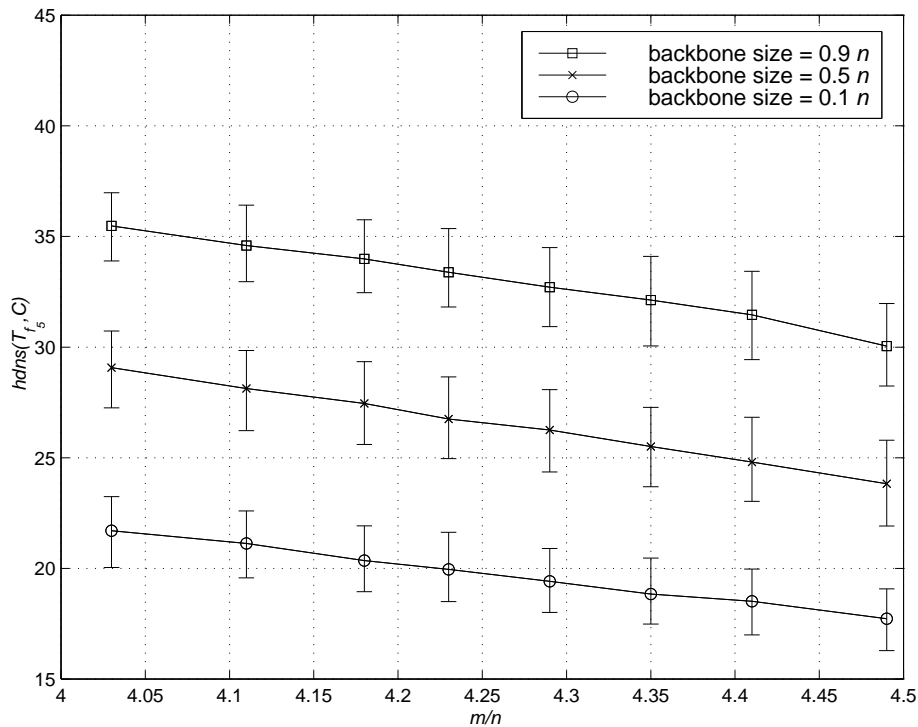

Figure 8: The effect of varying $m/n$ on $hdns(T_{f_5}, C)$ while backbone size is controlled.

Figure 8 showed that while backbone size is controlled, $hdns(T_{f_5}, C)$ falls linearly as $m/n$ is increased. The gradient of the fall is about $-14$. Table 2 showed that if backbone size is controlled and $m/n$ fixed, $hdns(T_{f_5}, C)$ is linearly related to log of cost, with the gradient of the fit being around 0.08. Given that this linear relationship continues to hold with a constant gradient as $m/n$ is varied (in fact the gradient decreases slightly) and assuming that increasing $m/n$ is not affecting the cost by other means, we would expect a linear decrease in log of mean cost with gradient $-1.12$, which is only slightly steeper than the observed decrease in log of median cost shown in Figure 4.

So the results are consistent with the idea that whatever factor causes the cost to decay exponentially as $m/n$ is varied does so largely by causing $hdns(T_{f_5}, C)$ to fall linearly.

## 3.3 Discussion

We have identified a pattern in search behaviour which is strongly related to the pattern in cost discussed in Section 2.3. Our interpretation of this pattern is as follows. In each instance the quasi-solutions which WSAT visits form interconnected areas of the search space such that local search can always move to a solution from them, without often moving to an assignment where many clauses are unsatisfied. The evidence for this is simply that WSAT runs are apparently always successful but visit the assignments where more clauses are unsatisfied very infrequently. Frank, Cheeseman and Stutz (1997) also mentioned in their analysis of GSAT search spaces that in Random 3-SAT, local minima where few clauses were unsatisfied can usually be escaped by unsatisfying just one clause.

| $m/n$ | Backbone size | Intercept of $lsr$ fit | Gradient of $lsr$ fit | $r$ |
|---|---|---|---|---|
| 4.03 | $0.1n$ | 1.0528 | 0.0844 | 0.9445 |
|      | $0.5n$ | 0.6928 | 0.0925 | 0.8769 |
|      | $0.9n$ | 0.7065 | 0.0895 | 0.7308 |
| 4.11 | $0.1n$ | 1.0166 | 0.0868 | 0.9511 |
|      | $0.5n$ | 0.6315 | 0.0955 | 0.8852 |
|      | $0.9n$ | 0.8158 | 0.0867 | 0.7196 |
| 4.18 | $0.1n$ | 1.0858 | 0.0839 | 0.9556 |
|      | $0.5n$ | 0.8090 | 0.0895 | 0.8799 |
|      | $0.9n$ | 0.8109 | 0.0864 | 0.7195 |
| 4.23 | $0.1n$ | 1.1290 | 0.0821 | 0.9581 |
|      | $0.5n$ | 0.8343 | 0.0887 | 0.8974 |
|      | $0.9n$ | 0.7480 | 0.0878 | 0.7691 |
| 4.29 | $0.1n$ | 1.1289 | 0.0826 | 0.9550 |
|      | $0.5n$ | 1.0032 | 0.0828 | 0.8935 |
|      | $0.9n$ | 0.8382 | 0.0856 | 0.7579 |
| 4.35 | $0.1n$ | 1.1664 | 0.0811 | 0.9628 |
|      | $0.5n$ | 0.9835 | 0.0842 | 0.8996 |
|      | $0.9n$ | 0.9835 | 0.0808 | 0.7728 |
| 4.41 | $0.1n$ | 1.2029 | 0.0795 | 0.9565 |
|      | $0.5n$ | 1.0274 | 0.0830 | 0.9135 |
|      | $0.9n$ | 1.1070 | 0.0768 | 0.7816 |
| 4.49 | $0.1n$ | 1.2458 | 0.0777 | 0.9661 |
|      | $0.5n$ | 1.1472 | 0.0787 | 0.9197 |
|      | $0.9n$ | 1.1930 | 0.0742 | 0.8086 |

Table 2: Data on correlations between $hdns(T_{f_5}, C)$ and $\log_{10}$ cost with $n = 100$ and $m/n$ and backbone size fixed at different values.

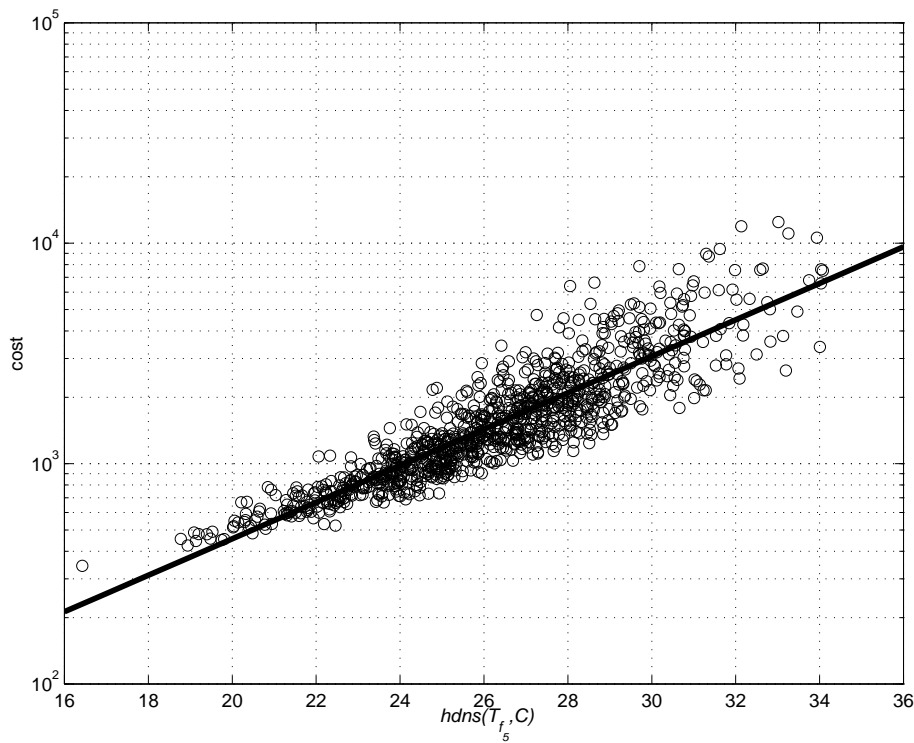

Figure 9: The relationship of $hdns(T_{f_5}, C)$ to log of cost when backbone size is controlled at $0.5n$ and $m/n$ is fixed at 4.29.

We believe that in instances of higher cost this quasi-solution area extends to parts of the search space which are Hamming-distant from solutions, whereas in instances of lower cost the area is less extensive. The mean Hamming distance between the early quasi-solution $T_{f_5}$ and the nearest solution is an accurate indicator of how extensive the quasi-solution area is. This interpretation suggests why $hdns(T_{f_5}, C)$ is so strongly correlated with cost: the extensiveness of the quasi-solution area determines how costly it is to search. It also suggests why, on instances of higher cost, quasi-solutions are found slightly more quickly: when the quasi-solution area is extensive, from a random starting point a shorter series of hill-climbing flips is required to find a quasi-solution.

The mean $hdns(T_{f_5}, C)$ decreases linearly as $m/n$ is increased while backbone size is controlled. At the same time, cost decays exponentially. We think this is because as $m/n$ is increased, the quasi-solution area becomes progressively less extensive.

## 4. A Causal Hypothesis

The pattern in search behaviour from Section 3 and our interpretation of it suggested a causal hypothesis to account for the decay in cost discussed in Section 2.3 and hence the overall peak. The key to this hypothesis is a property of SAT instances: backbone fragility. This property is qualitatively consistent with the above observations. Most importantly, although backbone fragility has implications for an instance's search space topology, it is a property based on the logical structure of the SAT instance. In this section we motivate and define backbone fragility, discuss how it may be measured and show how it relates to the patterns reported in Sections 2.3 and 3.

### 4.1 Backbone Fragility : Motivation

Suppose $B$ is a small sub-bag of the clauses of a satisfiable SAT instance $C$, such that there exists a set of quasi-solutions $Q_B$ where at most the clauses $B$ are unsatisfied. What structural property of $C$ would cause the quasi-solutions $Q_B$ to be attractive to WSAT? We already know that if the backbone of a Random 3-SAT instance is small, its solutions are found with little search (Parkes, 1997). The solutions to $C - B$ ($C - B$ denotes $C$ with one copy of each member of $B$ removed) are either solutions to $C$ or members of $Q_B$. If we assume that the assignments which are attractive to WSAT on $C$ are approximately the same assignments which are attractive on $C - B$, then the members of $Q_B$ (which are solutions of $C - B$) will be attractive in $C$ when the backbone of $C - B$ is small, particularly if $C$'s backbone is large. Furthermore for any $T_B \in Q_B$, the number of variables which do not appear in the backbone of $C - B$ is an upper bound on $hdns(T_B, C)$, so a large reduction in the backbone size allows for high $hdns(T_B, C)$. To summarise, if the removal of a certain small sub-bag of clauses causes the backbone size to be greatly reduced, we can expect that quasi-solutions where only these clauses are unsatisfied will be attractive to WSAT and possibly Hamming-distant from the nearest solution.

We are interested in quasi-solutions in general rather than those in $Q_B$. If removing a random small set of clauses on average causes a large reduction in the backbone size, we say that the instance is *backbone-fragile*. Where the effect on the backbone is smaller on average, the instance is *backbone-robust*. If a large-backbone instance is backbone-fragile, by extension of the above argument we expect that in general quasi-solutions will be attractive and they may be Hamming-distant from the nearest solution. Hence this idea is consistent with our observations and interpretation in Section 3: backbone fragility approximately corresponds to how extensive the quasi-solution area is.

The idea that backbone fragility is the underlying factor causing the search behaviour pattern is appealing for other reasons. For each entailed literal $l$ of $C$, there must be a sub-bag of clauses in $C$ whose conjunction entails $l$. For any given backbone size, as clauses are added, for any given entailed literal $l$ we expect that the extra clauses allow alternative combinations of clauses which entail $l$. Hence after adding clauses whilst controlling the backbone size, the random removal of clauses will have less effect on the backbone since the fact that a literal is entailed depends less on the presence of particular sub-bags. As clauses are added, we expect that instances will become less backbone-fragile. Given the hypothetical relationship between backbone fragility and the search behaviour, this would then explain qualitatively why the search behaviour changes as it does when $m/n$ is varied. We think that because backbone fragility is a property of the instance's logical structure, its study may also lead to further results about complexity issues, but we postpone discussion of this until Section 8.

## 4.2 The Measurement of Backbone Robustness

We now define a measure of the backbone robustness of an instance which will allow us to test predictions of the hypothesis. We take the instance $C$ and delete clauses at random, halting the process when the backbone size is reduced by at least half. At this point we record as the result the number of deleted clauses. This constitutes one *robustness trial*. Our metric for backbone robustness is the mean result of all such possible trials, i.e. the average number of random deletions of clauses which must be made so as to reduce the backbone size by half.

It is infeasible to compute the results of all possible robustness trials. Therefore, when measuring backbone robustness of an instance we estimated it by computing the average of a random sample of trials. We used at least 100 robustness trials in each case and in order to ensure a reasonably accurate estimate, we continued to sample more robustness trials until the standard error was less than $0.05 \times$ the sample mean (in which case our estimate of the mean was accurate to within about 10% at the 95% confidence level). With $n = 100$, using satisfiable instances from near the satisfiability threshold whose backbone size was controlled at 50, usually less than 250 robustness trials were required for the estimate to converge in this way. Even then, backbone robustness was costly to compute.

There were different possible metrics for backbone fragility/robustness, but we found that the metric described above gave the clearest results for our purposes without an unnecessarily complicated definition. Other metrics, such as the reduction in backbone size when a random fixed fraction of clauses are removed, may be more suitable in other contexts.

## 4.3 The Change in Backbone Robustness as the Control Parameter is Varied

As discussed in Section 4.1 we expect that if backbone size is controlled, backbone robustness increases as $m/n$ is increased. Since our measure of backbone robustness is defined in terms of the size of the backbone, it is most useful when comparing instances of equal backbone size.

We found that increasing the control parameter made instances more backbone-robust, as expected. Figure 10 shows the effect on backbone robustness of increasing $m/n$ through the satisfiability threshold while $n = 100$ and backbone size is controlled. Each point is the median of 1000 instances.

We note that backbone robustness as defined by our measure is generally higher for instances with larger backbones. We think that this is because on the large-backbone instances, the backbone must be reduced by a larger number of literals in each fragility trial and that this requires more clauses to be removed.

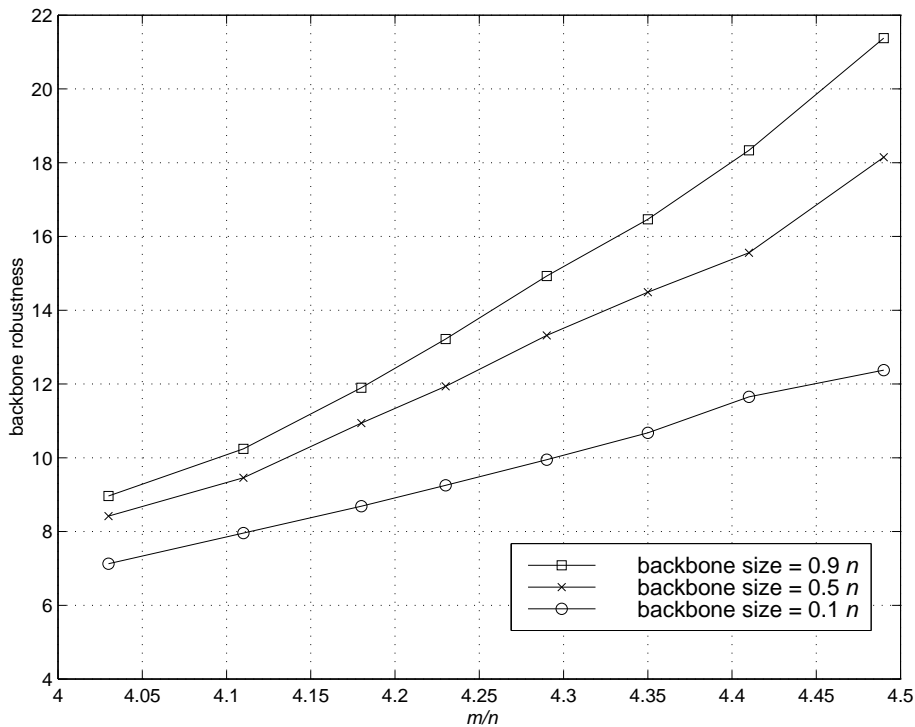

Figure 10: Backbone robustness through the satisfiability transition, with backbone size fixed at $0.1n$ $0.5n$ and $0.9n$.

## 5. A Correct Prediction about Cost Variance

We may assert that the fall of cost observed with the increase in the control parameter is due to the change in some other factor $F$, as for example Yokoo (1997) has. Such an assertion makes an important and testable prediction: that any variation in $F$ when the control parameter is fixed accounts for some of the variation in cost. However there may be other factors whose influence on the cost is so great as to obscure the effect of $F$ when the control parameter is fixed. To best reveal the effect of $F$, if there is any, the effects of some other factors may have to be controlled for.

Backbone robustness is our proposed factor $F$. The backbone size is another factor which strongly influences the cost. Our result in this section is that when the effects of $m/n$ and backbone size are controlled for, i.e. when they are fixed, the effects of backbone robustness can be seen quite clearly for large-backbone instances.

### 5.1 Correlation Data

Figure 11 shows a plot of the log cost against the measure of backbone robustness for Random 3-SAT instances with $n = 100$, $m/n$ 4.29 and backbone size controlled at $0.1n$, $0.5n$ and $0.9n$. A linear *lsr* fit is superimposed in each case. Table 3 gives the intercept, gradient and $r$ values for *lsr* fits with backbone size controlled at the same three values and with $m/n$ varied through the threshold.

| $m/n$ | Backbone size | Intercept of *lsr* fit | Gradient of *lsr* fit | $r$ | $r^{-95\%}$ | $r^{+95\%}$ | Rank corr. coefficient |
|---|---|---|---|---|---|---|---|
| 4.03 | $0.1n$ | 3.0338 | $-0.0204$ | $-0.1928$ | $-0.2506$ | $-0.1400$ | $-0.1934$ |
|  | $0.5n$ | 3.7075 | $-0.0370$ | $-0.3730$ | $-0.4191$ | $-0.3235$ | $-0.3713$ |
|  | $0.9n$ | 4.2846 | $-0.0419$ | $-0.4711$ | $-0.5165$ | $-0.4251$ | $-0.4699$ |
| 4.11 | $0.1n$ | 2.9639 | $-0.0134$ | $-0.1490$ | $-0.2088$ | $-0.0873$ | $-0.1402$ |
|  | $0.5n$ | 3.6675 | $-0.0351$ | $-0.3891$ | $-0.4355$ | $-0.3417$ | $-0.3770$ |
|  | $0.9n$ | 4.2287 | $-0.0370$ | $-0.4535$ | $-0.5001$ | $-0.4065$ | $-0.4662$ |
| 4.18 | $0.1n$ | 2.9365 | $-0.0146$ | $-0.1745$ | $-0.2356$ | $-0.1149$ | $-0.1663$ |
|  | $0.5n$ | 3.6067 | $-0.0302$ | $-0.3840$ | $-0.4272$ | $-0.3389$ | $-0.3738$ |
|  | $0.9n$ | 4.1811 | $-0.0338$ | $-0.5306$ | $-0.5687$ | $-0.4921$ | $-0.5466$ |
| 4.23 | $0.1n$ | 2.9257 | $-0.0155$ | $-0.2107$ | $-0.2659$ | $-0.1553$ | $-0.2116$ |
|  | $0.5n$ | 3.5142 | $-0.0239$ | $-0.3643$ | $-0.4105$ | $-0.3154$ | $-0.3436$ |
|  | $0.9n$ | 4.1313 | $-0.0312$ | $-0.5253$ | $-0.5645$ | $-0.4818$ | $-0.5457$ |
| 4.29 | $0.1n$ | 2.8766 | $-0.0136$ | $-0.1894$ | $-0.2483$ | $-0.1300$ | $-0.2053$ |
|  | $0.5n$ | 3.4903 | $-0.0225$ | $-0.3863$ | $-0.4350$ | $-0.3395$ | $-0.3996$ |
|  | $0.9n$ | 4.0934 | $-0.0290$ | $-0.5325$ | $-0.5721$ | $-0.4931$ | $-0.5467$ |
| 4.35 | $0.1n$ | 2.8261 | $-0.0109$ | $-0.1671$ | $-0.2250$ | $-0.1105$ | $-0.1724$ |
|  | $0.5n$ | 3.4325 | $-0.0199$ | $-0.3734$ | $-0.4222$ | $-0.3244$ | $-0.3782$ |
|  | $0.9n$ | 3.9939 | $-0.0237$ | $-0.4984$ | $-0.5394$ | $-0.4555$ | $-0.5243$ |
| 4.41 | $0.1n$ | 2.7925 | $-0.0100$ | $-0.1763$ | $-0.2321$ | $-0.1175$ | $-0.1683$ |
|  | $0.5n$ | 3.3772 | $-0.0172$ | $-0.3452$ | $-0.3954$ | $-0.2919$ | $-0.3582$ |
|  | $0.9n$ | 3.9284 | $-0.0211$ | $-0.5152$ | $-0.5582$ | $-0.4692$ | $-0.5270$ |
| 4.49 | $0.1n$ | 2.7164 | $-0.0073$ | $-0.1392$ | $-0.1926$ | $-0.0841$ | $-0.1355$ |
|  | $0.5n$ | 3.3506 | $-0.0170$ | $-0.4034$ | $-0.4585$ | $-0.3516$ | $-0.4027$ |
|  | $0.9n$ | 3.8720 | $-0.0198$ | $-0.5549$ | $-0.5949$ | $-0.5125$ | $-0.5604$ |

Table 3: Data on the correlation between backbone robustness and $\log_{10}$ cost with $n = 100$ and $m/n$ and backbone size fixed at different values.

The $r$ values suggest an effect of backbone robustness on cost, particularly for large backbone sizes. For smaller backbone sizes, we imagine that finding the backbone is less of an issue and so backbone fragility, which hinders this, has less of an effect. For the larger backbone sizes, we think the main difficulty for WSAT is satisfying the backbone; backbone fragility is then important. However, given the somewhat unclear shape of the scatter plots, there are several concerns as to the significance of the correlation, which we now address using some simple statistical methods.

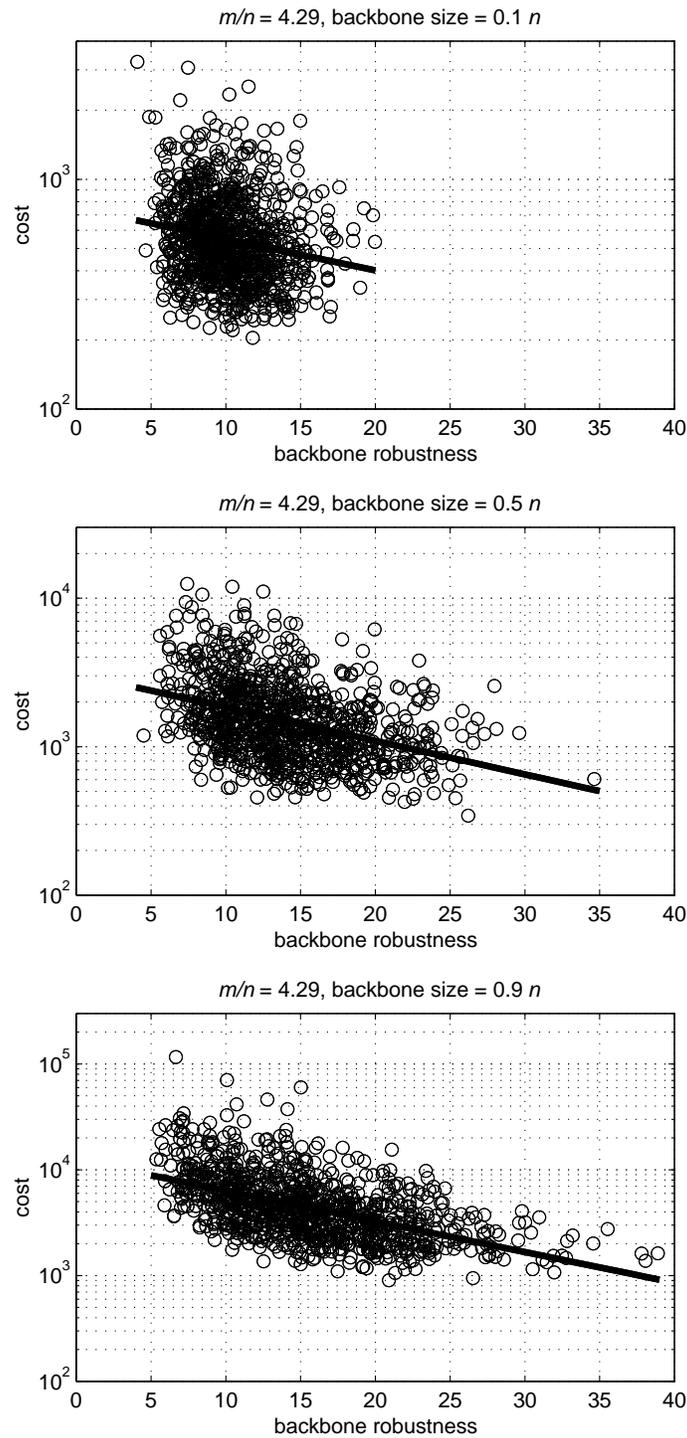

Figure 11: Scatter plot of backbone robustness and cost with $n = 100$, $m/n = 4.29$ and backbone size fixed at $0.1n$, $0.5n$ and $0.9n$.

## 5.2 An Artifact of the Distributions of the Variables?

One concern is that the observed $r$ could also have arisen simply because of the distributions of the two variables rather than because of any relationship between them. This is a serious concern here as the distributions are unknown.

The null hypothesis, $H_0$ is that the value of $r$ which results from the distributions of the two variables is equal to the observed $r$. A randomisation method can be used to test $H_0$. See Appendix A for details of this method. For each data set presented, 1000 randomised pairings of the data were constructed. In each case, we found that the observed $r$ does not fall within the range of the sampling distribution of $r$ for randomised pairings. $H_0$ can therefore be rejected at the 99.9% confidence level.

The $r$ coefficient, given above, can be greatly affected by outliers. Therefore the rank correlation coefficient, which is less affected, was also calculated. The rank correlation is also given in Table 3. We found that in each case the rank correlation coefficient is not considerably different from the $r$ coefficient. This demonstrates that the observed $r$ was not greatly affected by outliers.

## 5.3 Confidence Intervals for the Correlation

Given that there is a relationship between the two variables which is not merely an artifact of the distributions or of outliers, how accurate is our measurement of $r$? A bootstrap method can be used to obtain bounds on a confidence interval for this statistic. Again, the reader is referred to Appendix A for details of this method. Using this method with 1000 pseudo-samples we obtained lower and upper bounds on the 95% confidence interval for $r$, which are also given in Table 3 as $r^{-95\%}$ and $r^{+95\%}$ respectively. The data implies with 95% confidence, the upper bounds on the amount of error in our estimates of $r^2$ are between about 0.02 and 0.05.

## 6. A Correct Prediction about Very Backbone-Fragile Instances

Our hypothesis proposes that high backbone fragility of instances quite accurately represents the factor which (via the search behaviour patterns uncovered in Section 3) causes high WSAT cost for those instances. However, it is plausible that the high backbone fragility is a by-product of some unmeasured latent factor and that it is not causally related to the cost.

To help establish the causal link between backbone fragility and cost, we therefore created sets of random SAT instances which had higher backbone fragility than usual Random 3-SAT instances. This is to some degree following the methodological precedent of Bayardo and Schrag (1996), who created random instances which contained small unsatisfiable sub-instances but which had few constraints overall. These were often found to be exceptionally hard for the complete procedure NTAB. Their experiments thereby helped establish that this feature of instance structure was the cause of exceptionally high cost for complete procedures.

We cannot easily set backbone fragility directly, since it is not a generation parameter. One manipulation experiment which is possible is the use of an instance generation procedure which results in instances with a higher backbone fragility. Our hypothesis predicts that instances generated using such a procedure will be harder than Random 3-SAT instances. In this section we define such a procedure and test the prediction. It may be that our procedure is also manipulating the latent factor. However, since the procedure is specifically designed to increase backbone fragility, a correct prediction here still lends credibility to our hypothesis.

## 6.1 Backbone-minimal Sub-instances

Suppose we have a SAT instance $C$ and we remove a clause such that the backbone is not affected by the removal of the clause. If such clauses are repeatedly removed, eventually the instance will be such that no clause can be removed without disturbing the backbone. In this case we have a *backbone-minimal sub-instance* (BMS) of $C$. More formally, we have the following definition:

**Definition** A SAT instance $C'$ is a BMS of $C$ iff

- $C'$ is a sub-instance of $C$ (i.e. a sub-bag of the clauses) such that $C'$ has the same backbone as $C$.
- for each clause $c$ of $C'$ there exists a literal $l$ such that:
    1. $C' \rightarrow l$
    2. $(C' - \{c\}) \wedge \neg l$ is satisfiable

  i.e. every strict sub-instance of $C'$ has a strictly smaller backbone than the backbone of $C'$ □

BMSs can be seen as satisfiable analogues of the *minimal unsatisfiable sub-instances* (MUSs) of unsatisfiable instances studied by amongst others Culberson and Gent (1999b) in the context of graph colouring and Gent and Walsh (1996) and Bayardo and Schrag (1996) in satisfiability. An MUS of an instance $C$ is a sub-instance which is unsatisfiable, but such that the removal of any one clause from the sub-instance renders it satisfiable. Just as all unsatisfiable instances must have an MUS, all satisfiable SAT instances must have a BMS. Having a BMS does not depend on having a non-empty backbone – if the backbone of the instance is empty, its BMS is the empty sub-instance. An instance can have more than one BMS. Different BMSs of an instance may share clauses. One BMS of an instance cannot be a strict sub-instance of another.

Suppose the backbone of a satisfiable instance $C$ is the set of literals $\{l_1, l_2, \ldots, l_k\}$. Let $d$ be the clause $\neg l_1 \vee \neg l_2 \vee \ldots \vee \neg l_k$. Then we have the following useful fact:

**Theorem** $C'$ is a BMS of $C$ iff $C' \wedge d$ is an MUS of $C \wedge d$ □

A simple proof of the above is given in Appendix B. Due to this fact, methods for studying MUSs can be applied to the study of BMSs. We can study the BMSs of a satisfiable instance $C$ by finding the backbone of $C$ and then studying the MUSs of $C \wedge d$: each of these corresponds to a BMS of $C$ since $d$ must be present in every MUS of $C \wedge d$.

To find a BMS of $C$ we determine the backbone, then find a random MUS of $C \wedge d$ using the same MUS-finding method as Gent and Walsh (1996) and remove $d$ from the result.

| Instances | Backbone robustness | | |
|---|---|---|---|
| | 10th percentile | Median | 90th percentile |
| PRESERVE-BACKBONE($C$, 0, $C'$) | 8.5845 | 12.9700 | 20.6300 |
| PRESERVE-BACKBONE($C$, 5, $C'$) | 7.7374 | 12.0317 | 19.1700 |
| PRESERVE-BACKBONE($C$, 10, $C'$) | 7.1977 | 11.0851 | 17.3500 |
| PRESERVE-BACKBONE($C$, 20, $C'$) | 6.0690 | 9.3913 | 14.5351 |
| PRESERVE-BACKBONE($C$, 40, $C'$) | 4.2622 | 6.4899 | 9.9900 |
| PRESERVE-BACKBONE($C$, 80, $C'$) | 2.0745 | 2.8661 | 3.9851 |
| BMS | 1.0200 | 1.0600 | 1.1600 |

Table 4: The effect of PRESERVE-BACKBONE on backbone robustness.

## 6.2 Interpolating Between an Instance and one of its BMSs

Once a BMS $C'$ has been established, we can also study the effects of interpolation between $C$ and $C'$ by removing at random from $C$ some of the clauses which do not appear in $C'$. This is equivalent to removing clauses at random such that the backbone is preserved. PRESERVE-BACKBONE($C$, $m_r$, $C'$) will denote $C$ with $m_r$ clauses, which do not appear in the BMS $C'$, removed at random. The resulting instance will have the same backbone as $C$.

Just as increasing $m/n$ while controlling the backbone size causes backbone robustness to increase, we have found that deleting clauses such that the backbone is unaffected causes backbone robustness (as measured above) to decrease, as one might expect.

We used 500 Random 3-SAT instances with $n = 100$ and $m/n = 4.29$. For each instance we found one BMS. We then used PRESERVE-BACKBONE to interpolate with $m_r$ set at various values. Table 4 shows the effect of increasing $m_r$ on backbone robustness. The BMSs of the threshold instances are so backbone-fragile that the removal of just one clause from them is likely to reduce the backbone by a half or more.

Our hypothesis predicts that as this interpolation from $C$ to $C'$ proceeds, the cost for local search increases because the backbone robustness decreases. It is conceivable, although it would be very surprising, that removing *any* clauses from random instances near the threshold generally makes their cost for local search increase. If this were the case, any increase in cost during interpolation towards a BMS could merely be due to the removal of clauses *per se* rather than the removal of clauses whilst preserving the backbone. To control for this possibility we also removed clauses according to two other procedures. The procedure RANDOM($C$, $m_r$) removes $m_r$ clauses from $C$ at random. The procedure REDUCE-BACKBONE($C$, $m_r$) removes $m_r$ clauses such that each time a clause is removed, the size of the backbone is reduced. The clause to be removed is chosen randomly from all such clauses. This procedure therefore uses the opposite removal criterion to PRESERVE-BACKBONE. If the backbone becomes empty, no further clauses are removed.

Figure 12 shows the effect on per-instance cost of applying the three clause removal procedures to the same set of 500 Random 3-SAT threshold instances. Each plot is the median per-instance cost.

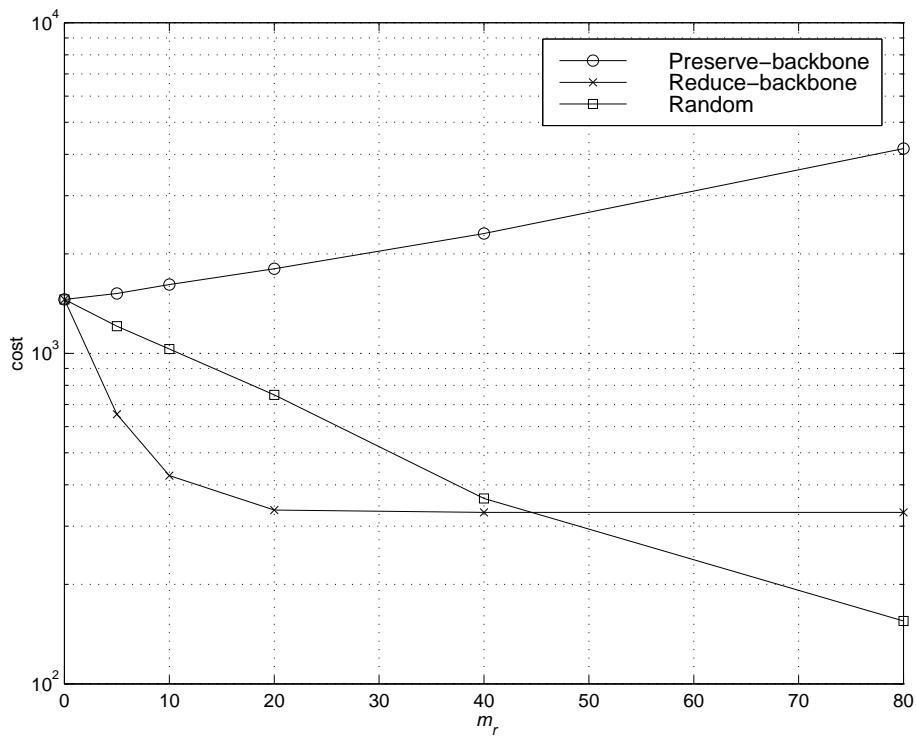

Figure 12: The effect of the three clause removal procedures on median per-instance cost.

We observe that removing clauses randomly or such that the backbone is strictly reduced, causes cost to be reduced, so the removal of clauses does not in itself cause higher cost. The REDUCE-BACKBONE procedure causes a greater initial fall in cost, as the backbone size is reduced more quickly than with RANDOM. However, the cost then stabilises for REDUCE-BACKBONE because the backbone becomes empty and thereafter no more clauses are removed.

Removing clauses according to PRESERVE-BACKBONE causes the local search cost to increase by an amount approximately exponential in the number of clauses removed. Table 5 gives more data on this effect and also cost data for BMSs. The interpolation shifts the whole distribution up, not just the median. The median cost of the BMSs, which are the most backbone-fragile of all the instances, is more than three times that of the 90th cost percentile of Random 3-SAT instances.

The BMSs of these instances had between 254 and 318 clauses. The above results therefore demonstrate the existence of instances in the underconstrained region which are much harder than the typical instances from near the satisfiability threshold. However since these were not obtained by sampling from Random 3-SAT directly, we do not know how often they occur. As far as we know, they are vanishingly rare and therefore, in contrast to exceptionally hard instances for complete algorithms, it seems unlikely that they affect the mean cost. Also, while Gent and Walsh (1996) showed that the exceptionally hard instances for complete algorithms are hard for a different reason from that of threshold instances, BMSs are apparently hard for the same reason – because they are backbone-fragile.

| Instances | Per-instance cost | | |
|---|---|---|---|
| | 10th percentile | Median | 90th percentile |
| PRESERVE-BACKBONE($C$, 0, $C'$) | 517 | 1450 | 5175 |
| PRESERVE-BACKBONE($C$, 5, $C'$) | 537 | 1515 | 5657 |
| PRESERVE-BACKBONE($C$, 10, $C'$) | 557 | 1608 | 6009 |
| PRESERVE-BACKBONE($C$, 20, $C'$) | 570 | 1803 | 7037 |
| PRESERVE-BACKBONE($C$, 40, $C'$) | 643 | 2295 | 10683 |
| PRESERVE-BACKBONE($C$, 80, $C'$) | 816 | 4154 | 24313 |
| BMS | 1556 | 16945 | 135883 |

Table 5: The effect of PRESERVE-BACKBONE on per-instance cost.

One useful by-product of this section is a means of generating harder test instances for local search variants without increasing $n$. However these instances do require $O(m + n)$ complete searches to generate: $O(n)$ to determine satisfiability and the backbone and $O(m)$ to reduce to a BMS.

## 7. A Correct Prediction about Search Behaviour

Recall that in the motivating discussion of Section 4.1 it was suggested that the quasi-solutions in $Q_B$ would be attractive if the backbone of $C - B$ was small. That is to say that the clauses of $B$ are more likely to be the set of unsatisfied clauses if the removal of the clauses of $B$ has a large effect on the backbone. This part of the hypothesis also makes a prediction about search behaviour – that clauses most often unsatisfied by WSAT should be those whose removal reduces the backbone size most. In this section we show this prediction to be correct.

We looked at individual instances which were cost percentiles from a set of 5000 Random $k$-SAT instances with $n = 100$ and $m/n = 4.29$. Per-instance cost was determined as in previous sections. For each clause in the instance, we calculated the number of backbone literals which were no longer entailed if the clause was removed. This is a simple measure of the *backbone contribution* ($bc$) of the clause – how much the backbone size depends on the presence of the clauses. If a clause's backbone contribution is high, it is termed a *backbone-critical* clause. We made 1000 runs of WSAT on each instance under the same conditions as in previous sections. During search, each time the current assignment changed we recorded whether each clause was unsatisfied. The result of averaging the number of times the clause was unsatisfied over all runs gives the *unsatisfaction frequency* ($uf$) of that clause.

Figure 13 shows a plot of these two quantities for the clauses of the instance whose cost was median of 5000 threshold instances. We note from this figure that the clauses whose presence contributes the most to the backbone are more often unsatisfied than average during WSAT search.

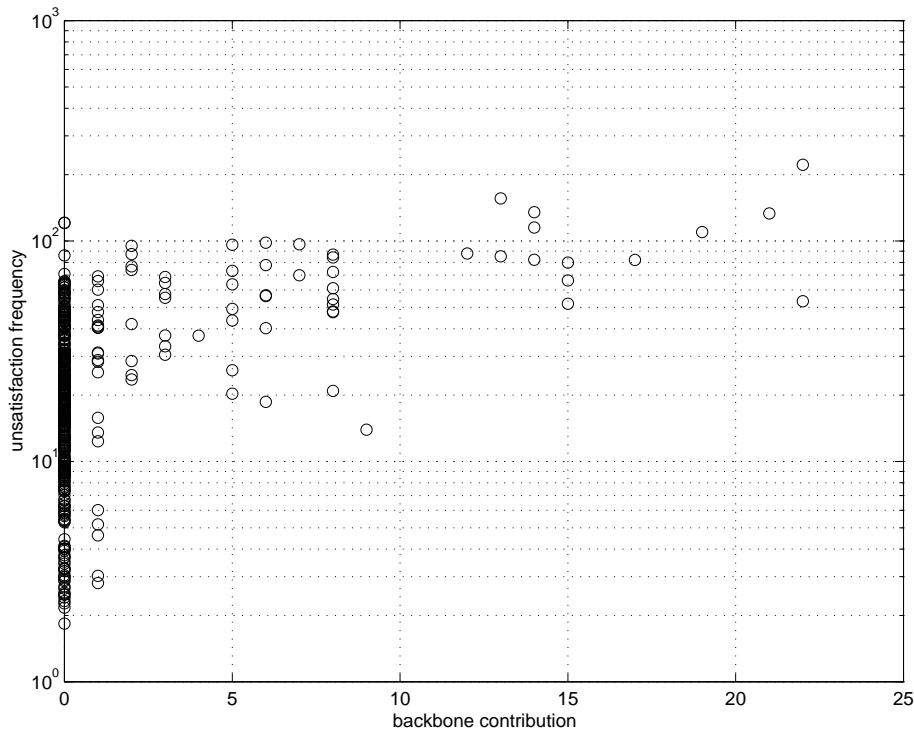

Figure 13: Scatter plot of unsatisfaction frequency against backbone contribution for the clauses of the cost median of 5000 instances, $m/n = 4.29$, $n = 100$.

Table 6 confirms this pattern. Each row of the table gives data for one instance. We selected cost percentiles; individual instances of varying degrees of difficulty. For example the row labelled '30th' corresponds to the instance whose cost is the 1500th in rank from the easiest to the most difficult of the 5000 instances, while the 50th percentile instance is the one used to produce Figure 13. The third and fourth columns give the mean and standard deviation of the unsatisfaction frequency over all clauses in the instance and the last two columns give the same statistics for the sub-bag of the clauses which were most backbone-critical (their backbone contribution was in the top 10%).

Table 7 shows that the converse effect is also present: the clauses which are most often unsatisfied (their unsatisfaction frequency is in the top 10%) are more backbone-critical than average. Although an effect is quite clear from the means, there are sometimes particularly large standard deviations in $bc$ values for the most frequently unsatisfied clauses. This is because, as can be seen from Figure 13, some clauses are very often unsatisfied even though removing them on their own does not affect the backbone size at all. We have found in other experiments that the removal of these clauses along with other small random bags of clauses does on average reduce the backbone size considerably. The large standard deviations therefore arise because the true backbone contribution of these clauses is not apparent when using this simple measure.

| Cost Percentile | Backbone size | All clauses | | Most backbone-critical clauses | |
|---|---|---|---|---|---|
| | | $uf$ mean | $uf$ std. dev. | $uf$ mean | $uf$ std. dev. |
| 10th | 16 | 11.3430 | 8.0704 | 20.8703 | 8.8730 |
| 20th | 11 | 13.1079 | 10.9596 | 30.1817 | 16.8730 |
| 30th | 13 | 21.0207 | 16.3680 | 41.4660 | 21.1142 |
| 40th | 36 | 22.9825 | 21.3118 | 56.6841 | 27.4660 |
| 50th | 48 | 29.5615 | 25.9275 | 72.0704 | 38.7779 |
| 60th | 25 | 36.2940 | 35.6327 | 96.1664 | 54.3081 |
| 70th | 63 | 52.4198 | 48.1078 | 119.7691 | 66.8187 |
| 80th | 70 | 92.2623 | 87.7827 | 167.3428 | 149.8058 |
| 90th | 93 | 108.3124 | 127.1968 | 306.7200 | 198.6933 |

Table 6: Unsatisfaction frequencies of clauses in different cost percentile instances.

| Cost Percentile | Backbone size | All clauses | | Most often unsatisfied clauses | |
|---|---|---|---|---|---|
| | | $bc$ mean | $bc$ std. dev. | $bc$ mean | $bc$ std. dev. |
| 10th | 16 | 0.5921 | 1.2358 | 2.0909 | 1.8529 |
| 20th | 11 | 0.4848 | 1.0380 | 1.7727 | 1.9632 |
| 30th | 13 | 0.3963 | 1.2405 | 1.8409 | 2.4490 |
| 40th | 36 | 1.8089 | 4.2411 | 8.3182 | 6.4620 |
| 50th | 48 | 1.0629 | 3.2781 | 6.3182 | 6.6043 |
| 60th | 25 | 1.3800 | 3.4920 | 7.7500 | 5.7794 |
| 70th | 63 | 3.3916 | 8.8630 | 14.9091 | 15.8126 |
| 80th | 70 | 0.6946 | 3.4577 | 2.5909 | 7.8602 |
| 90th | 93 | 3.0653 | 10.0376 | 16.9318 | 20.6436 |

Table 7: Backbone contributions of clauses in different cost percentile instances.

For instances of different costs at the satisfiability threshold, the clauses which are most likely to be unsatisfied during search have a higher backbone contribution than average. Conversely, the clauses which have the largest backbone contribution are more likely to be unsatisfied during search. This section therefore demonstrates that as well as explaining differences in cost between instances, the backbone fragility hypothesis can also explain differences in the difficulty of satisfying particular clauses during search.

## 8. Related and Further Work

Clark *et al.* (1996) showed that the number of solutions is correlated with search cost for a number of local search algorithms on random instances of different constraint problems, including Random 3-SAT. The pattern was confirmed by Hoos (1998) using an improved methodology. Clark *et al.*'s work was the first step towards understanding the variance in cost when the number of constraints is fixed. We have followed their approach both by looking at the number of solutions and by using linear regression to estimate strengths of relationships between factors.

Schrag and Crawford (1996) made an early empirical study of the clauses (including literals) which were entailed by Random 3-SAT instances. Parkes (1997), whose study is also discussed in Section 1, looked in detail at backbone size in Random 3-SAT and its effect on local search cost. He also linked the position of the cost peak to that of the satisfiability threshold by the emergence of large-backbone instances which occurs at that point. Parkes also identified the fall in WSAT cost for instances of a given backbone size. This was therefore the basis for our study. Parkes conjectured that the presence of a "failed cluster" may be the cause of high WSAT cost for some large-backbone Random 3-SAT instances. According to this hypothesis, the addition of a single clause could remove a group of solutions which is Hamming distant from the remaining solutions, reducing the size of the backbone dramatically. Such a clause would then have a large backbone contribution. Therefore our explanation for the general high cost of the threshold region has certain features in common with Parkes' conjecture. In particular we agree that it is the presence of clauses with a large backbone contribution which causes high cost. This is especially demonstrated by our results from Section 7.

Frank *et al.* (1997) studied in detail the topology of the GSAT search space induced by different classes of random SAT instances. Their study discussed the implications of search space structure for future algorithms, as well as the effects of these structures on algorithms such as GSAT. They also noted that some local search algorithms such as WSAT may be blind to the structures they studied because they search in different ways to GSAT.

Yokoo (1997) also addressed the question of why there is a cost peak for local search as $m/n$ is increased. The approach was to analyse the whole search space of small satisfiable random instances. While in this study, we have only examined SAT, Yokoo also showed his results generalised to the colourability problem. Yokoo used a deterministic hill-climbing algorithm. He studied the number of assignments from which a solution is reachable (solution-reachable assignments) via the algorithm's deterministic moves, which largely determines the cost for the algorithm.

We followed Yokoo in looking for a factor competing with the number of solutions whose effect on cost changes as $m/n$ is increased. The factor which Yokoo proposed as the cause of the overall fall in cost was the decrease in the number of local minima – assignments from which no local move decreased the number of unsatisfied clauses. The decrease in this number was demonstrated as $m/n$ is increased. The decrease was attributed to the decreasing size of "basins" (interconnected regions of local minima with the same number of unsatisfied clauses). Yokoo claimed (p. 363) that:

> "adding constraints [...] makes the [instance] easier by decreasing the number of local minima".

However, we do not think it is clear *a priori* what the relationship between the number of local minima and the cost is in a given instance and Yokoo did not study it sufficiently to convince us of his explanation. In contrast with Yokoo, we have studied in detail the relationship between the backbone fragility of instances and WSAT's cost on these instances and confirmed it by testing predictions of our hypothesis. Also, we studied instance properties that related to the logical structure of the clauses rather than the search space topology which was induced as we think this has more potential to generalise across algorithms and even to address complexity issues, as we explain towards the end of this section.

Hoos (1998) also analysed the search spaces of SAT instances in relation to local search cost by looking at two new measures of the induced objective function which he defined, including one based on local minima. Although via these measures, Hoos was not able to account for the Random 3-SAT cost peak, he found that the features were correlated with cost for some SAT encodings of other problems and has also shown (Hoos, 1999b) that his measures can help distinguish between alternative encodings of the same problem.

How does the pattern we have uncovered fit in to other work on what makes instances require a high cost to solve? Gent and Walsh (1996) looked at the probability that an unsatisfiable SAT instance became satisfiable if a fixed number of clauses are removed at random. The unsatisfiable instances which had the highest computational cost for a complete procedure were found to be those which were unsatisfiability-fragile – their unsatisfiability was sensitive to the random removal of clauses. It may therefore be that the fragility of an instance's unsatisfiability or backbone size is the cause of high computational cost both in the context of complete procedures and incomplete local search, which would be an interesting link between the two algorithm classes. This link may form the basis of a possible explanation of the reasons why threshold Random 3-SAT instances may be universally *hard* in the average case, as opposed to merely costly for some class of algorithms. Recent work by Monasson *et al.* (1999a, 1999b) has suggested that parameterised distributions of instances which are hard in the average case, e.g. Random 3-SAT, exhibit a discontinuity in the backbone size[3] as the control parameter is varied, whereas in polynomial time average-case distributions, such as Random 2-SAT, the backbone size changes smoothly. They propose that the complexity of the distribution is linked to the presence of this discontinuity. We conjecture that this may be because in the asymptotic limit, instances which are backbone- or unsatisfiability-fragile only persist as $n$ is increased where there is such a discontinuity. This line of research may therefore establish a testable causal mechanism for this pattern, showing *how* the properties of the instance distributions affect algorithm performance.

It would be interesting to compare backbone fragility in different random distributions of 3-SAT instances, such as those introduced by Bayardo and Schrag (1996) and by Iwama,

---

3. Monasson *et al.*'s definition of the backbone also extends to unsatisfiable instances.

Miyano and Asahiro (1996) to see whether differences in local search cost could be explained. A method which generates satisfiable instances which are quickly solved by local search is analysed by Koutsoupias and Papadimitriou (1992) and Gent (1998). Random clauses are added to the formula as in Random 3-SAT but only if they do not conflict with a certain solution which is set in advance. We conjecture that overconstrained examples of these are quickly solved by local search because they are very backbone-robust.

An interesting possibility mentioned by Hoos and Stützle (1998) suggested by the exponential run length distribution, was that local search is equivalent to random generate-and-test in a drastically reduced search space. We conjecture that this reduced search space corresponds to the quasi-solution area. Measurements of $hdns(T_B, C)$ for quasi-solutions $T_B$ may therefore be indicative of the extensiveness of this reduced search space, especially since this metric is linearly correlated with log cost. Further experimentation in this vein may therefore reveal more about the topology of the reduced search space which could in turn lead to better local search algorithms designed to exploit this knowledge.

Finally, we should emphasise that the notions of backbone and backbone-fragility are equally applicable to non-random SAT instances. In future we may be able to confirm that the results we have shown for random SAT instances apply equally to benchmark and real-world SAT instances. However, one caveat here is that entailed literals may be uncommon in these instances and we may need to study the fragility of other sets of entailed formulas.

## 9. Conclusion

We have reconsidered the question of why cost for local search peaks near the Random 3-SAT satisfiability threshold. The overall pattern is one of two competing factors. The cause of the onset of high cost as the control parameter is increased has been previously established as the decreasing number of solutions. We have proposed that the cause of the subsequent fall in cost is falling backbone fragility.

We found a striking pattern in the search behaviour of the local search algorithm WSAT. For instances of a given backbone size, in the underconstrained region of the control parameter, WSAT is attracted early on to quasi-solutions which are Hamming-distant from the nearest solution. This distance is also very strongly related to search cost. As the control parameter is increased, the distance decreases. We suggested backbone fragility was the cause of this pattern.

We defined a measure of backbone robustness. Backbone-fragile instances have low robustness. We were then able to test predictions of the hypothesis that the fall in backbone fragility is the cause of the overall decay in cost as the control parameter is increased. We found that the hypothesis made three correct predictions. Firstly that the degree to which an instance is backbone-fragile is correlated with the cost when the effects of other factors are controlled for. Secondly, that when Random 3-SAT instances are altered so as to be more backbone-fragile (by removing clauses without disrupting the backbone) their cost increases. Thirdly, that the clauses most often unsatisfied during search are those whose deletion has most effect on the backbone.

We now summarise our interpretation of the evidence. In the underconstrained region, instances with small backbones are predominant. In this region, the rapid hill-climbing phase typically results in an assignment which is close to the nearest solution (and probably satisfies the backbone). Since finding the small backbone is largely accomplished by hill-climbing, typical cost for WSat is low in this region and variance in cost is due to variance in the density of solutions in the region of the search space where the backbone is satisfied.

In the threshold region, large-backbone instances quickly appear in large quantities. For large-backbone instances, the main difficulty for local search is to identify the backbone rather than to find a solution once the backbone has been identified. The identification of a large backbone may be accomplished by the rapid hill-climbing phase to a greater or lesser extent. We think that this extent is determined by the backbone fragility of the instance. If a large-backbone instance is backbone-fragile the hill-climbing phase is ineffective and results in an assignment which is Hamming-distant from the nearest solution (probably implying that much of the backbone has not been identified). Then a costly plateau search is required to find a solution. Hence when the rare large-backbone instances do occur in the underconstrained region, they are extremely costly to solve because of their high backbone fragility.

If a large-backbone instance is more backbone robust, the rapid hill-climbing phase is more effective in determining the backbone and the plateau phase is shorter. So overall the instance is less costly for WSat to solve. Hence for large-backbone instances, since backbone fragility increases as we add clauses, cost decreases. In the overconstrained region, large backbone instances are dominant and so backbone fragility becomes the main factor determining cost. Hence cost decreases in this region. Our hypothesis proposes the following explanation for the cost peak: *Typical cost peaks in the threshold region because of the appearance of many large-backbone instances which are still moderately backbone-fragile, followed by the increasing backbone robustness of these instances.*

## Acknowledgments

This research was supported by UK Engineering and Physical Sciences Research Council studentship 97305799 to the first author. The first two authors are members of the cross-university Apes research group (`http://www.cs.strath.ac.uk/~apes/`). We would like to thank the other members of the Apes group, the anonymous reviewers of this and an earlier paper and Andrew Tuson for invaluable comments and discussions.

## Appendix A: Randomisation and Bootstrap Tests

We summarise the methods as used in this context. Further explanation of these methods is given in Cohen (1995).

### A.1 Randomisation for Estimating the Correlation Coefficient due to the Distributions of the Variables

Randomisation can be used to estimate the correlation coefficient between the two variables which results simply from their distributions rather than from any relationship. We start with the two vectors of data $\bar{x} = \langle x_1, x_2, \ldots, x_N \rangle$ and $\bar{y} = \langle y_1, y_2, \ldots, y_N \rangle$. If the correlation coefficient is merely due to the distributions of $x$ and $y$, then it is not dependent on any particular $x_i$ being paired with $y_i$. Therefore to calculate the correlation coefficient resulting merely from the distributions we pair the $x$ and $y$ data randomly.

We construct $K$ randomisations. Each randomisation consists of a vector $\bar{y}'$, which is simply a random permutation of $\bar{y}$. For each randomisation, we calculate the correlation coefficient between $\bar{x}$ and $\bar{y}'$ – note that each value $x_i$ is now paired with a random value from $\bar{y}$. These randomised correlation coefficients give us an estimate of the correlation coefficients resulting from the distributions of the variables. If $K$ is large enough, we will have an accurate estimate which can be compared with the correlation coefficient in the observed data.

### A.2 Bootstrap Estimation of Confidence Intervals for Correlation Coefficients

We have an original sample $\langle (x_1, y_1), (x_2, y_2), \ldots (x_N, y_N) \rangle$ of $N$ pairs. A *pseudo-sample* from the original also consists of $N$ pairs. The $j^{\text{th}}$ pair in the pseudo-sample $(x_j^b, y_j^b) = (x_q, y_q)$ where $q$ is a random number between 1 and $N$. Each pair in the pseudo-sample is chosen independently i.e. pairs are sampled from the original with replacement. We assume that our original sample of pairs of data is representative of the whole population of such pairs. Given this, composing pseudo-samples is just like sampling from the whole population. Therefore by measuring the correlation coefficient of many pseudo-samples, we can study what the correlation coefficient would have looked like had we taken many sets of samples from the whole population. From the distribution of the correlation coefficient among many pseudo-samples (the bootstrap sampling distribution) we can infer bounds on the confidence interval for the observed correlation coefficients.

Many pseudo-samples are taken, and the correlation coefficient is calculated for each of the pseudo-samples. This gives the bootstrap sampling distribution of the correlation coefficient. The $97.5^{\text{th}}$ percentile of this distribution is an upper bound on the 95% confidence interval for the correlation coefficient, and the $2.5^{\text{th}}$ percentile is a lower bound.

## Appendix B: The Relationship Between BMSs and MUSs

Let $C$ be a satisfiable SAT instance and $\{l_1, l_2, \ldots, l_k\}$ be the set of all literals entailed by $C$. Let $d$ be the clause $\neg l_1 \lor \neg l_2 \lor \ldots \lor \neg l_k$.

**Theorem** $C'$ is a BMS of $C$ iff $C' \land d$ is an MUS of $C \land d$ $\square$

PROOF Suppose $C'$ is a BMS of $C$. Then $C' \wedge d$, which is a sub-instance of $C \wedge d$, must be unsatisfiable, as $d$ violates every literal in the backbone of $C'$. If $d$ is removed from $C' \wedge d$, the result $C'$ is satisfiable. If any other clause $c$ is removed from $C' \wedge d$, there must be some literal from the backbone of $C'$, $l_i$ say, such that $(C' - \{c\}) \wedge \neg l_i$ is satisfiable. Therefore, since $\neg l_i$ is also a literal of $d$, $(C' - \{c\}) \wedge d$ is satisfiable. Therefore $C' \wedge d$ is an MUS of $C \wedge d$.

Conversely, suppose $C' \wedge d$ is an MUS of $C \wedge d$. Since $C' \wedge d$ is minimally unsatisfiable, $C'$ is satisfiable. Since $C'$ is a sub-instance of $C$, the backbone of $C'$ must be a subset of the backbone of $C$. Suppose there were some literal $l_j$ which was in the backbone of $C$ but not in the backbone of $C'$. Then there would be a solution to $C' \wedge \neg l_j$. This would then also be a solution to $C' \wedge d$, since $\neg l_j$ is one literal of $d$. This contradicts $C' \wedge d$ being unsatisfiable and so there can be no $l_j$ i.e. $C'$ and $C$ must have the same backbone.

$C' \wedge d$ is minimally unsatisfiable. Therefore for any clause $c$ of $C'$, $(C' - \{c\}) \wedge d$ is satisfiable. Any solution to $(C' - \{c\}) \wedge d$ must make some literal $\neg l_k$ of $d$ true, and must therefore also be a solution to $(C' - \{c\}) \wedge \neg l_k$. Therefore $l_k$, which is in the backbone of $C'$, is not in the backbone of $(C' - \{c\})$. Hence $C'$ is a BMS of $C$ □